\documentclass[10pt,twocolumn,letterpaper]{article}

\usepackage{cvpr}
\usepackage{times}
\usepackage{epsfig}
\usepackage{graphicx}
\usepackage{amsmath}
\usepackage{amssymb}
\usepackage{enumitem}

\usepackage{subfigure}

\usepackage{pifont}
\newcommand{\no}{\ding{55}}
\newcommand{\yes}{\checkmark}

\usepackage[colorlinks]{hyperref}

\cvprfinalcopy 

\ifcvprfinal\fi
\begin{document}

\title{Moviescope: Large-scale Analysis of Movies using Multiple Modalities}

\author{Paola Cascante-Bonilla$^1$\thanks{Indicates equal author contribution.} \quad Kalpathy Sitaraman$^2$\thanks{Work was conducted while affiliated with the University of Virginia.} \footnotemark[1]  \quad Mengjia Luo$^1$ \quad Vicente Ordonez$^1$\\
$^1$University of Virginia, $^2$Microsoft\\
{\tt\small [pc9za, ml6uk, vicente]@virginia.edu, kasivara@microsoft.com}}

\maketitle

\begin{abstract}
   Film media is a rich form of artistic expression. Unlike photography, and short videos, movies contain a storyline that is deliberately complex and intricate in order to engage its audience. 
   In this paper we present a large scale study comparing the effectiveness of visual, audio, text, and metadata-based features for predicting high-level information about movies such as their genre or estimated budget. We demonstrate the usefulness of content-based methods in this domain in contrast to human-based and metadata-based predictions in the era of deep learning. Additionally, we provide a comprehensive study of temporal feature aggregation methods for representing video and text and find that simple pooling operations are effective in this domain. We also show to what extent different modalities are complementary to each other. 
    To this end, we also introduce Moviescope, a new large-scale dataset of 5,000 movies with corresponding movie trailers (video + audio), movie posters (images), movie plots (text), and metadata. 
\end{abstract}

\section{Introduction}

As recording equipment, and internet speeds become a commodity, both independent and professional video production and distribution become possible. In fact, popular movie streaming services such as Hulu, Netflix, or Amazon Video occupy today a sizeable amount of the bandwidth of the entire Internet\footnote{Netflix Accounts for More than A Third of All Internet Traffic. Time Business. May 2015. http://time.com/3901378/netflix-internet-traffic/}. Thus, designing and evaluating representations for automatic understanding of this type of data becomes increasingly important. Movies unlike short videos provide a narrative of a story that requires a holistic understanding of a long range of events that can be depicted both as a sequence of images and sounds (video), or a sequence of words (text). In addition, movies are a popular medium of artistic expression, and reveal our preferences and potentially insightful aspects of our culture, so their analysis is also of great intrinsic interest.  

\begin{figure}[t]
\centering
\includegraphics[width=0.98\linewidth]{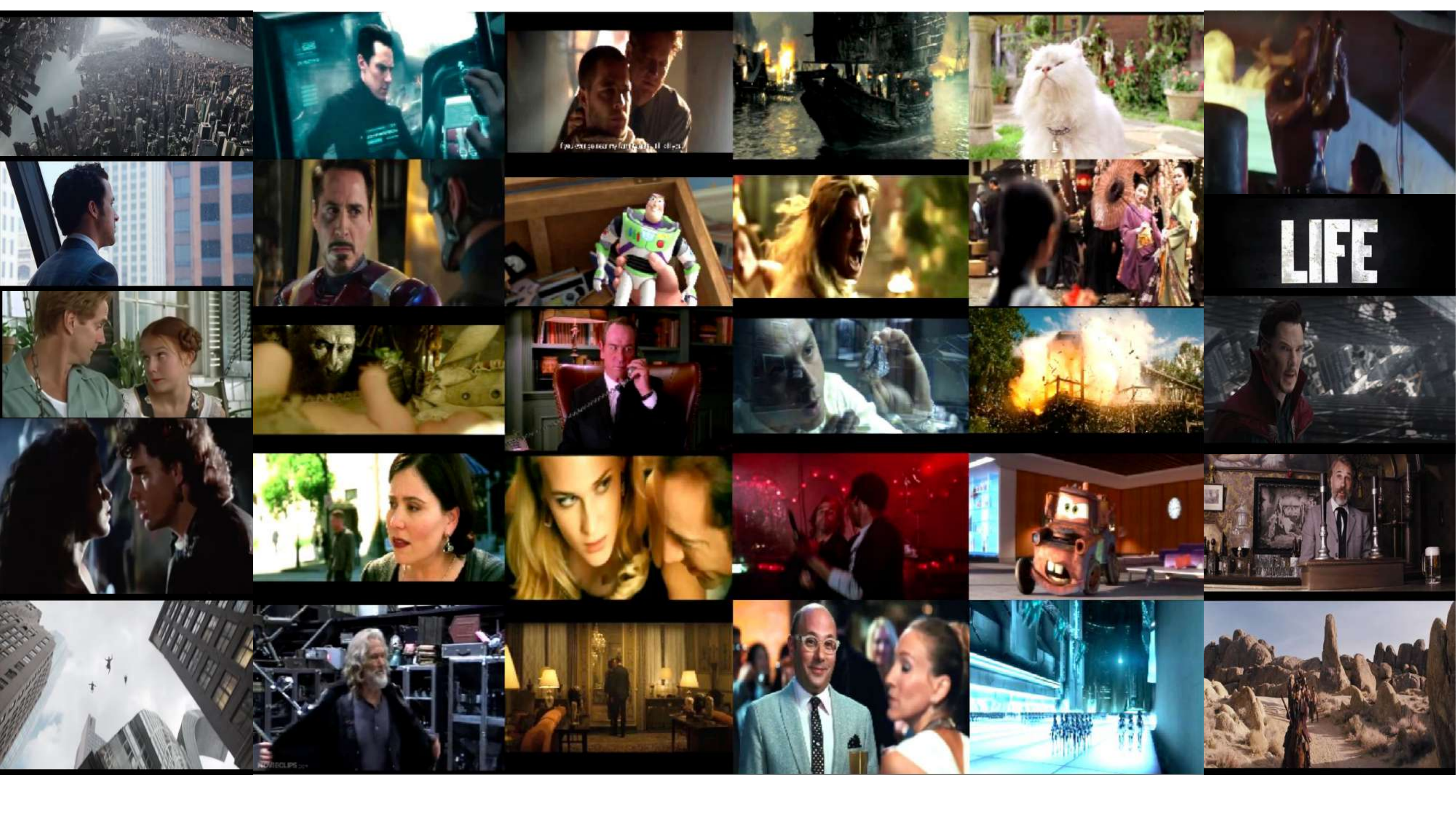}
\vspace{-0.05in}
\caption{Sample frames from Moviescope video trailers showing the diversity in the proposed dataset. For each of the 5,000 video trailers we have also compiled corresponding movie posters, movie plots, and previoulsy collected metadata from IMDB.}
\vspace{-0.15in}
\label{fig:lead}
\end{figure}

There have been significant efforts in the community to compile video datasets for an increasing number of semantic concepts \cite{soomro2012ucf101,heilbron2015activitynet,sigurdsson2016hollywood,abu2016youtube}. However most of previous efforts have focused on short clips depicting isolated activities. Less effort has been dedicated on building video resources that capture long temporal semantics as exemplified by movies. One limitation has been practical;  processing long hours of video would require extremely large amounts of computation. In our work, we propose to use video trailers to represent each movie. Video trailers bring two important practical aspects: they are a compromise between short clips spanning a few seconds and full-length movies spanning a few hours each, and video trailers are publicly available for most movies. These two modalities, along with movie posters, are compiled for a diverse set of 5,000 movies, therefore constituting a rather complete new resource for multimodal analysis. We plan to release our crawled plots and video trailers (in the form of urls to publicly available video), manually curated mappings between these two modalities, pre-trained audio, video, poster, and plot embeddings, as well as code to reproduce our experiments. We believe that novel and diverse datasets such as Moviescope will greatly enhance the ability of the community to advance research on different aspects of movie understanding. We show a sample of frames from video trailers in Figure~\ref{fig:lead}.

We also present an extensive study for the use of different modalities in predicting high-level attributes for movies and report key findings. For instance, we demonstrate that video trailers are able to capture sufficient evidence of their corresponding full-length movies to make predictions about movie genre, thus are --to some degree-- a reasonable summary of the movie for this purpose. Similarly, we show that movie plots collected from Wikipedia for the same movies are also a good predictor for movie genres but provide complimentary information. We also find that audio from a trailer is a better predictor of the budget of the movie than the corresponding video from the trailer.  
We use \emph{fastText}~\cite{fasttext} for representing movie plots, and propose an analogous encoding mechanism for video, which we refer as \emph{fastVideo}. These two encodings use simple average pooling operations over word-level and frame-level features. We show the promise of these simple encoding mechanisms for video as well as text, and show that they compare favorably against other more sophisticated methods such as recurrent neural networks. 

Our contributions can be summarized as follows:
\begin{itemize}[noitemsep,topsep=1pt,parsep=3pt,partopsep=0pt]
\item We introduce Moviescope, a multimodal movie dataset of video+audio trailers, text plots, movie posters, and metadata associated with more than 5,000 movies, which includes crawled data and manually curated mappings between these modalities.
\item We evaluate fastText~\cite{fasttext} and an equivalent representation we name \emph{fastVideo} for encoding video frames using a time pooling operations and compare against other feature aggregation approaches, and prior work.
\item We present a detailed benchmark of various multimodal encodings based on text, video, audio, posters and metadata for the task of movie genre prediction and budget estimation. Including a user study comparing our models against human performance for movie genre prediction using several modalities.
\end{itemize}

\section{Related Work}
There has been considerable effort in building video-centric datasets for action recognition, especially for actions involving people such as KTH~\cite{schuldt2004recognizing}, HDMB51~\cite{Kuehne11}, UCF50~\cite{reddy2013recognizing}, UCF101~\cite{soomro2012ucf101}, THUMOS~\cite{idrees2017thumos}, and Sports1M~\cite{karpathy2014large}. These datasets have been widely used to advance the field and have had a tremendous impact in the community. Movie trailers are considerably longer than the clips in these datasets, e.g.~UCF101 clips are around seven seconds long on average, while video trailers in Moviescope are on average two minutes long. Moreover, the videos are different because each trailer portrays many atomic actions, and often fantasy worlds and scenarios. 

More recent datasets include longer videos because they either depict activities composed of many actions~\cite{heilbron2015activitynet}, or they depict a sequence of actions~\cite{sigurdsson2016hollywood}. ActivityNet~\cite{heilbron2015activitynet} contains longer activities such as dancing, ironing clothes, fixing bicycle, etc. Because this dataset contains these types of activities as opposed to simpler actions, a sizeable proportion of the videos are between five and ten minutes long. The Charades dataset~\cite{sigurdsson2016hollywood}, unlike most previously described datasets, contains crowdsourced videos from online workers, as opposed to crawled video from Youtube, and the videos depict several actions, thus the videos are also longer than simple action datasets at thirty seconds on average. Movie trailers also depict many actions, or complex activities in the video but they are different from the previous datasets because they aim to represent a much longer series of events in the full movie which is hours of content.

There has also been previous work on videos from movies~\cite{cour2008movie,sankar2009subtitle,tapaswi2015book2movie,zhu2015aligning,tapaswi2016movieqa,rohrbach2017movie}, and TV series~\cite{Everingham2006HelloMN,sivic2009you,bauml2013semi,bojanowski2013finding,ramanathan2014linking}. The most common task among these works is a form of alignment between the videos and some other modality such as closed captions, movie scripts, audio descriptions~\cite{rohrbach2017movie}, or the corresponding book for the movie~\cite{tapaswi2015book2movie,zhu2015aligning}. These datasets are often limited in terms of number of movies because the tasks are designed to be within a movie, and not to make a holistic assessment of each movie as a data sample. We posit that by using video trailers (as opposed to full-length movies), and movie plots (as opposed to full-length movie scripts), we can find a compromise where this type of large scale analysis can be performed.

Finally, the closest related datasets in terms of specific domain and video type (movie trailers), are the one proposed by Zhou~et~al~\cite{MovieGenreViaSceneCat}, and the LTMD dataset introduced by Simoes~et~al~\cite{Simoes}. These two datasets also contain movie trailers and are focused on the task of movie genre prediction. We adopt this task as the canonical task in our dataset as well, and present detailed experiments for each modality and in combination. Beyond our dataset containing a larger amount of movie trailers, unlike these previous two datasets, our movie genres are more nuanced, containing thirteen non-mutually exclusive movie genre labels (multi-label classification), as opposed to four mutually exclusive movie genres (single-label classification) in~\cite{Simoes}. More importantly, Moviescope contains aligned movie plots (text), and movie posters (static images) for the same movies. We used a combination of automatic and manual curation to pair these plots with each movie trailer.


\section{Moviescope Dataset}

We introduce a novel dataset, Moviescope, which is based on the IMDb5000 dataset consisting of $5,043$ movie records. This dataset was released under an Open Database License as part of a Kaggle Competition, and contains a rich schema of metadata information about each movie including details about user interactions in social media. We significantly augmented this dataset by crawling video trailers associated with each movie from YouTube and text plots from Wikipedia. Next we describe in detail the methodology of our data collection.

\begin{figure}[t]
    \centering
    \subfigure[Genres co-occurrence heatmap]
    {
        \includegraphics[width=1.5in]{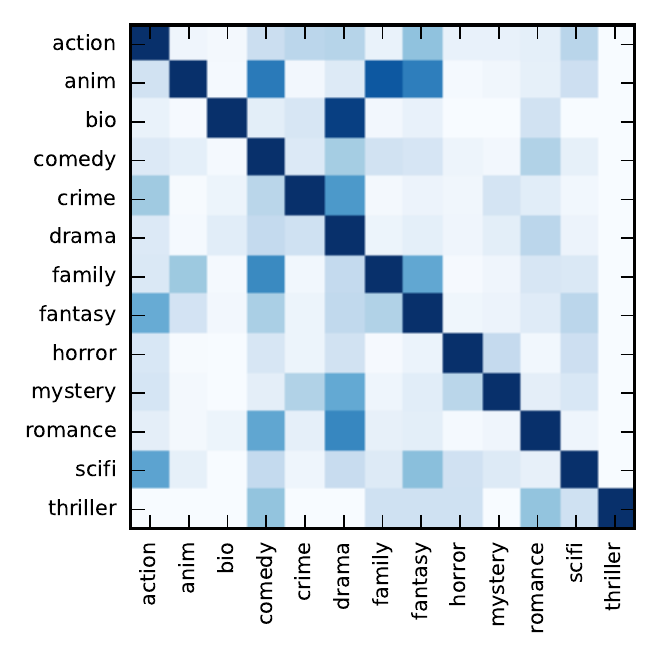}
        \label{fig:cooc}
    }
    \subfigure[Distribution of Genres]
    {
        \includegraphics[height=1.3in,width=1.5in]{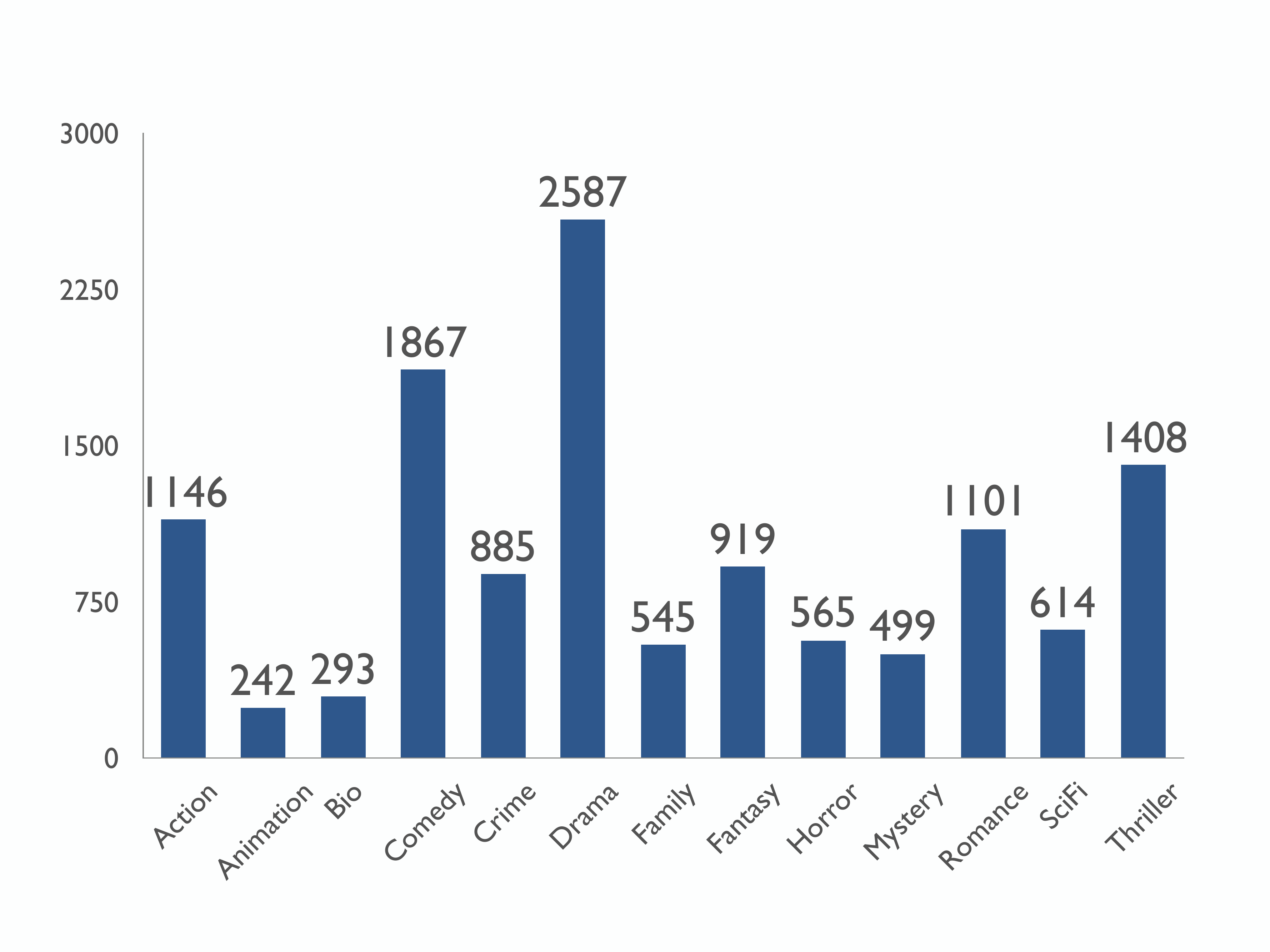}
        \label{fig:dist}
    }
    \caption{Label statistics in the Moviescope dataset, showing co-ocurrence and distribution of genres.}
    \label{fig:data-analysis}
    \vspace{-0.15in}
\end{figure}


\noindent{\bf Video trailers}: For each movie in the dataset, we search for its corresponding video trailer on YouTube using a script that fetches the first search result and downloads that particular video. 
We guarantee downloading the best possible video trailer by adding the term ``trailer'' at the end of the movie name on the automatically issued search query. 
We could successfully download videos for $4,996$ movies 
using this approach, 
and for the remaining movies -- most of which were short-films or movies from the golden era -- we downloaded their trailers manually. 

\vspace{0.04in}
\noindent{\bf Text plots}: We query for the contents of the Wikipedia page corresponding to each movie and crawl its contents whenever the page is available. 
The response HTML is then parsed using regular expressions to look for the plot of the movie, which is then preprocessed to remove unnecessary white-spaces and non-ASCII characters. 
If the movie page is not found on Wikipedia or the plot could not be parsed, we use the curated summaries from the CMU Movie Summary Corpus\cite{cmu}, which is another dataset of movies with associated text and metadata but no video. For movies that were not available in this dataset, we scrape the IMDb page for the movie to look for its corresponding storyline.

\vspace{0.04in}
\noindent {\bf Metadata}: The proposed dataset is very diverse and consists of movies from all over the world, spanning $66$ countries, and $48$ languages, including movies from $2,399$ different directors and $4,100$ different actors. The oldest movie in the dataset is from 1920 and the latest is from 2016. There are $13$ movie genre categories: \emph{action}, \emph{animation}, \emph{biography}, \emph{comedy}, \emph{crime}, \emph{drama}, \emph{family},   \emph{fantasy}, \emph{horror}, \emph{mystery}, \emph{romance}, \emph{sci-fi}, and \emph{thriller}. Compared to previous works~\cite{MovieGenreViaSceneCat,Simoes} we experiment with a larger number of categories and also perform multi-label as opposed to single-label classification. We show in Figure~\ref{fig:cooc} the co-occurrence matrix for various labels in our dataset. For instance we can observe that \emph{animation} movies are frequently also labeled as \emph{family} movies, but the opposite is less frequent. We also show in Figure~\ref{fig:dist} the raw counts for the number of movies assigned to each movie genre, the most frequent category is \emph{drama} with $2,587$ samples, and the least frequent category is \emph{animation} with $242$ samples. 

\begin{table}
\centering
\small
\caption{Comparison with similar movie-centric datasets}
\vspace{0.02in}
    \begin{tabular}{lcccccl}
    \hline
    Dataset&size &cats.& trailer&plot&metad.\\
    \hline
    Zhou et~al~\cite{MovieGenreViaSceneCat}& 1,239 & 4 & \yes & \no & \no\\
    LMTD~\cite{Simoes} &3,500 & 4 &\yes & \no & \no\\
    Moviescope (ours) & 5,027 & 13 & \yes & \yes & \yes\\
  \hline
\end{tabular}
\label{table:datasets}
\vspace{-0.1in}
\end{table}


To the best of our knowledge, Moviescope is the first movie-centric multimodal dataset that compiles together video trailers, textual plots, movie posters (static images), and movie metadata. Table~\ref{table:datasets} shows a comparison of Moviescope against previously collected datasets with movie trailers.  Movie trailers, movie plots, and metadata potentially contain separate but complementary information about the movie as we demonstrate through our experiments. Our dataset is larger and has richer annotations compared to these two previous datasets that also include movie trailers. Moviescope consists of approximately 195 hours of video (or around 702k seconds), and more than $20$ million frames. The average duration of the trailers is $129$ seconds. We found $49,107$ unique words across all plot summaries with the average number of words in the plots being $549$ and the longest plot containing $2656$ words. We expect that this compiled resource along with our experiments serve as the starting point to build more complex multimodal techniques and more robust representations.

\section{Modal Representations}
\begin{figure}[t]
\centering
\includegraphics[width=\linewidth]{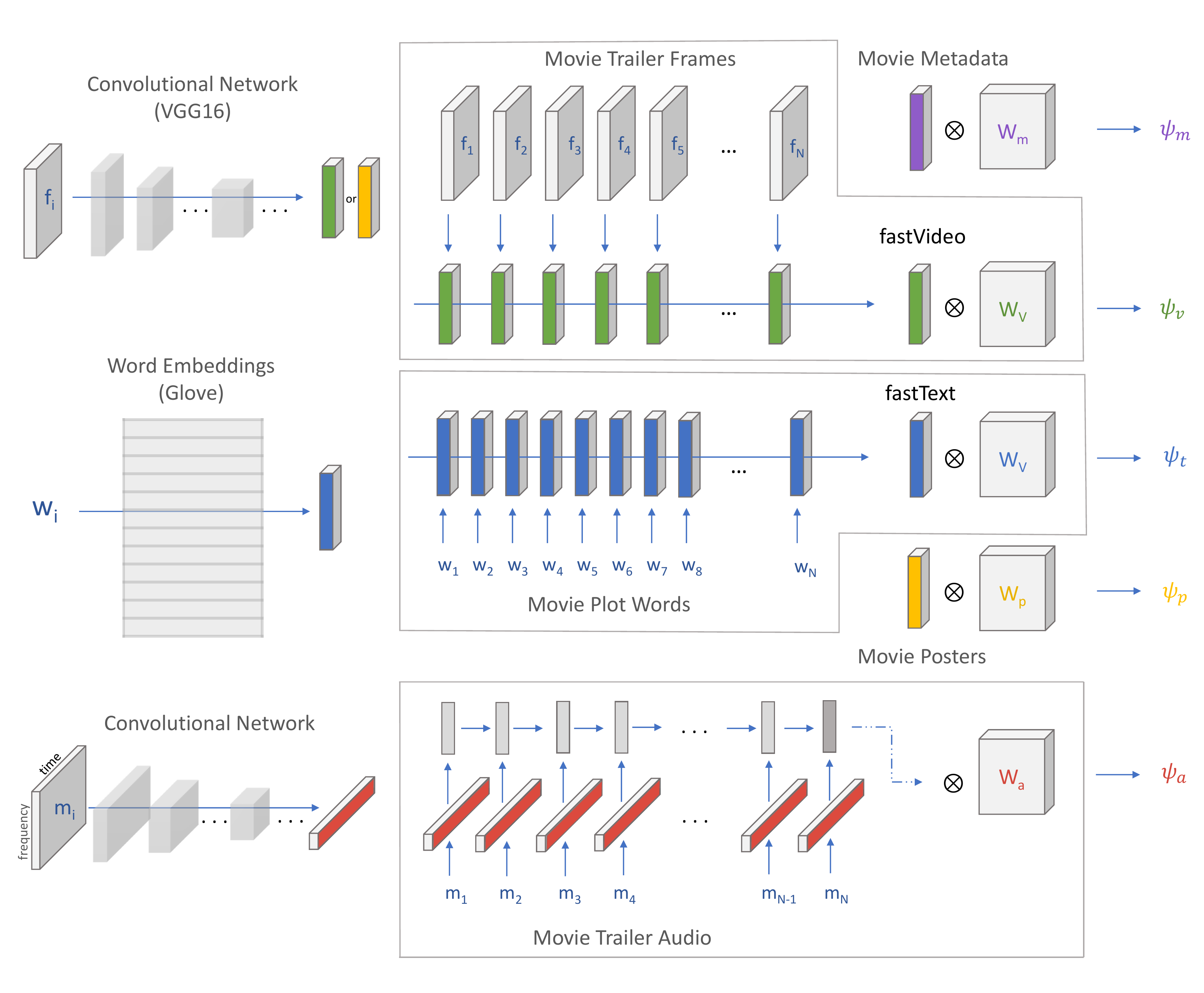}
\caption{Overview of our feature representations for multimodal movie encoding. On the left side we show how we extract a feature representation from a single frame using a CNN, a feature representation from a single word using word embeddings, and spectral-based features for audio representation, we then aggregate these features over time using a pooling operation. For metadata we use raw inputs with a random forest classifier.}
\label{fig:model}
\vspace{-0.1in}
\end{figure}

In this section, we discuss our feature representations for each individual modality: video, text, audio, posters and metadata. We use these representations for our task of movie genre prediction, and movie budget estimation but these can be applied to other similar tasks. In section~\ref{sec:fastText} we describe the text representation encoding mechanism used, which is inspired by the recently proposed fastText~\cite{fasttext} encoding. In this paper, the authors found that encodings based on simple time-pooling operations were competitive against encodings based on Long Short Term Memory networks (LSTMs)~\cite{lstm-original} for classification tasks. Inspired by this work we likewise use time-pooling operations for representing video, we call this encoding fastVideo (section~\ref{sec:fastVideo}). We show that this representation outperforms LSTM-based temporal feature aggregation. Sections~\ref{sec:audio} and ~\ref{sec:poster} describe the audio and poster representation mechanisms used. Finally, in section~\ref{sec:metadataRepresentation} we describe the types of entries in the movie metadata. Figure~\ref{fig:model} shows an overview and summary of our representations for each modality. 

\subsection{Text Representation - fastText}
\label{sec:fastText}
A common enconding for text relies on LSTM networks~\cite{lstm-original}, including bidirectional and stacked LSTM representations where the inputs are represented as a sequence of word embedding vectors. Word embeddings are continuous vector representations of words that are typically trained using an auxiliary task such as word context prediction through distributional semantics~\cite{word2vec,pennington2014glove}. Based on more recent research on textual representations for classification tasks (fastText\cite{fasttext}), we adopt this approach instead but also provide comparisons with LSTM-based encoders.
The fastText encoding mechanism works by computing a feature representation $\psi_t$ using a simple global average time-pooling operation over vector representations $w_j$ corresponding to unigrams (word embeddings), bigrams or trigrams. In the case of bigrams or trigrams, a temporal convolution layer is used to aggregate word embeddings among adjacent words. In the case of unigrams we would have: 
\begin{equation}
\psi_t = W_t \cdot \left[\frac{1}{N}\sum_{j=0}^{N}w_j\right]+b_t\, ,
\end{equation}
\noindent
where $N$ is the length of the sequence, and $W_t$,$b_t$ are the parameters of an affine layer that produces a task-specific number of outputs. In our unigram implementation of fastText, we encode the text in our movie plots using a fixed maximum length of 3000 words. We represent each word in this sequence using the Global Vector for Word Representations (GloVe) embeddings~\cite{pennington2014glove}. These embeddings were pretrained on text from Wikipedia with the unsupervised auxiliary task of predicting word context. We use the 300-dimensional version of the pretrained GloVe vectors. We also conduct experiments where we represent our words with randomly initialized word embeddings, and show that GloVe vectors provided some gain.

\subsection{Video Representation - fastVideo}
\label{sec:fastVideo}
Similarly as in our fastText implementation where an average time-pooling operation is performed over word embeddings, we propose fastVideo where the same operation is performed over the outputs of a convolutional neural network applied to individual video frames. We take 200 video frames subsampled by processing one frame every 10 frames. If the processed frames are less than the required amount, we complement it with more frames taken from the video starting at the 200th frame by taking one every 6 frames.
Subsequently, we input these frames into a pretrained VGG-16 convolutional neural network~\cite{DBLP:journals/corr/SimonyanZ14a} pretrained on Imagenet and use the activations from the penultimate layer of this model. This results in a feature encoding vector $f_i$ of size 4096. The time pooling operation works similarly as in fastText, where we aggregate either individual frame embeddings or frame embeddings corresponding to bigrams, or trigrams. In the case of bigrams we use a temporal convolutional layer with a stride size of two to aggregate embeddings between pairs of adjacent frames. In the case of \emph{frame unigrams} we would have:
\begin{equation}
\psi_v=W_v \cdot \left[\frac{1}{M}\sum_{i=0}^{M}f_i\right] +b_v\, ,
\end{equation}

\noindent
where $M$ is the number of frames in the sequence, and $W_v$,$b_v$ are the parameters of an affine layer that produces a task-specific number of outputs. 
We also compare this approach with other video representations for action recognition such as C3D~\cite{c3d} and Two-Stream I3D models~\cite{i3d}, which explicitly leverage temporal dynamics. In our experiments we make sure that C3D and i3D both have access to the same frames as fastVideo, and we show that for movie genre and budget estimation, the simpler and more efficient fastVideo encoding provides higher levels of accuracy.


\subsection{Audio Representation}
\label{sec:audio}
We extract the audio from each movie trailer and compute the log-\emph{mel} scaled power spectrogram to represent the power spectral density of the sound in a log-frequency scale. We use a set of 4 continuous clips of 30 seconds from the beginning of each audio and downsample them to 12kHz. When the audio is less than 2 minutes, we extract the required amount of remaining clips randomly from any point of the audio sample. The number of mel-bins is 128 and the hop-size is 256, resulting in 4 matrices of shape 128x1407. We stack those representations and use them as the inputs of a Convolutional Recurrent Neural Network (CRNN) that consists in four layers of spatial convolutions followed by two LSTM layers. This implementation is inspired by~\cite{musicTagger}, who used their model for music classification while still controlling the number of parameters with respect to the performance and training time per sample. We simply use this output representation $f_a$ to compute a task-specific number of logit unnormalized output scores using an affine transformation: $\psi_a = W_a \cdot f_a + b_a$, where $W_a$ and $b_a$ are parameters of this transformation.

\subsection{Poster Representation}
\label{sec:poster}
For each poster image we simply used the same pretrained VGG-16 network \cite{DBLP:journals/corr/SimonyanZ14a} as in section~\ref{sec:fastVideo} to precompute the features from the penultimate layer to represent each poster with a 4096-dimensional vector, and compute prediction scores as $\psi_p = W_p \cdot f_p + b_p$, where $W_p$ and $b_p$ are the parameters of an affine transformation layer from input representation $f_p$ to a task-specific number of logit unnormalized output scores.

\subsection{Metadata Representation}
\label{sec:metadataRepresentation}
The IMDb5000 dataset includes $28$ metadata entries including movie genres. For our experiments we did not use all metadata entries such as the aspect ratio of the movie, or whether the movie was shot in Black/White or Color, as these might either be uninformative for high-level semantic tasks. The metadata also includes IMDb plot keywords, but we did not consider these as we are using similar but richer textual representations using movie plots. 
We list in Table~\ref{table:metadata} the metadata entries used in our experiments along with their data type, and possible values. Categorical features such as \emph{director}, \emph{language}, and \emph{content-rating} are numerically-encoded, and appended alongside numeric features such as the \emph{number of faces in the poster}, \emph{duration}, \emph{number of likes on facebook}. The words in the movie title are represented by averaging their word embeddings.
For our single-modality experiments we use this feature on top of a random forest classifier to obtain scores $\psi_m$ for movie genre prediction. We additionally run a comparison with XGBoost~\cite{chen2016xgboost}, another popular model for heterogeneous data.

\begin{table}[t]
  \centering
  \small
  \caption{Metadata entries used in our experiments}
  \vspace{0.05in}
  \setlength{\tabcolsep}{5pt}
  {
  \begin{tabular}{llr}
    \hline
    Metadata Entry & Type & Values\\
    \hline
    \texttt{cast\_total\_facebook\_likes} & integer & --\\
    \texttt{duration} & integer & in mins\\
    \texttt{facenumber\_in\_poster} & integer & --\\
    \texttt{director\_name} & categorical & 1-1858\\
    \texttt{actor\_1\_name} & categorical & 1 - 2879\\
    \texttt{actor\_2\_name} & categorical & 1 - 2879\\
    \texttt{actor\_3\_name} & categorical & 1 - 2879\\
    \texttt{language} & categorical & 1 - 48\\
    \texttt{movie\_title} & GloVe & 300-dim\\
    \texttt{num\_critic\_reviews} & integer & --\\
    \texttt{movie\_facebook\_likes} & integer & --\\
    \texttt{content\_rating} & categorical & 1 - 18\\
    \texttt{num\_voted\_users} & integer & --\\
  \hline
\end{tabular}
}
\label{table:metadata}
\end{table}

\subsection{Multimodal Fusion}


In order to combine multiple modalities, we use the  output scores from the models associated with each individual modality as inputs to a weighted regression in order to obtain final movie genre predictions. 
Given the five modalities $M = \{t,v,a,p,m\}$, we can compute weight scores $\alpha_{ji}$ for the $j^{th}$ category (e.g. movie genre) and for each $i^{th}$ modality $\in M$:

\begin{equation}
\alpha_{ji} = \frac{e^{W_{ji}}}{\sum_{i \in M} e^{W_{ji}}},
\end{equation}
where $W_{ij}$ are extra learned parameters to fuse the modalities. The final prediction scores $\psi$ are then calculated as:
\begin{equation}
\psi = \alpha_{v}\psi_{v} + \alpha_{a}\psi_{a} + \alpha_{p}\psi_{p} + \alpha_{t}\psi_{t} + \alpha_{m}\psi_{m}
\end{equation}
Intuitively, these weights in the linear combination can be interpreted as the contribution of each modality toward predicting a genre; and since both $\psi$ scores and $\alpha$ coefficients are calibrated, we could consider them as a form of \emph{modal attention}. 


\section{Experimental Settings}
First, we discuss various baseline models used for comparison against our proposed encoding models. We also provide additional details about training and dataset setup.

\begin{table*}[t]
  \caption{Mean Average Precision (mAP) Scores for movie genre prediction.}
  \vspace{0.02in}
  \label{tab:prc}
  \resizebox{\textwidth}{!}{
  \begin{tabular}{l|ccccccccccccc|ccc}
    \hline
    &action&anim&bio&
    com&crime&drama&fam&fant&horr&
    myst&rom&scifi&thrlr&$mAP$&$\mu AP$&$sAP$\\
    \hline
    \% of training samples&8.70&1.84&2.22&14.17&10.56&19.63&4.14&6.97&4.29&3.79&8.36&4.66&10.69&-&-&-\\
    Baseline accuracy&22.1&4.3&6.2&39.3&18.6&53.6&10.8&17.0&10.5&10.9&22.1&13.5&25.8&19.6&13.7&21.0\\
    \hline
    
    \textbf{Video (V)}\\
    C3D~\cite{c3d} &63.8&91.3&16.2&82.3&45.1&71.6&65.3&54.8&50.8&28.2&38.3&21.8&64.8&53.4&57.9&68.8\\
    I3D~\cite{i3d} &37.2&51.8&9.2&72.6&33.9&67.6&43.6&39.0&22.8&21.3&34.3&22.6&48.3&38.8&50.5&65.6\\
    LSTM &47.5&86.8&12.0&79.2&33.0&72.0&64.5&54.4&22.7&24.7&40.4&36.5&54.8&48.4&59.6&70.5\\
    Bidirectional LSTM &49.9&86.3&8.2&77.6&29.9&70.8&65.4&55.3&22.3&21.7&41.6&35.9&51.2&47.4&58.2&69.9\\
    fastVideo &61.4&94.8&23.9&81.5&41.7&77.0&67.0&62.6&36.1&30.4&48.4&48.2&62.0&56.5&64.9&75.6\\
    fastVideo + TempConv &64.7&95.7&21.2&83.5&49.1&78.9&68.6&68.9&42.7&29.2&46.8&51.0&64.8&\textbf{58.9}&\textbf{65.9}&\textbf{76.3}\\
    \hline
    \textbf{Audio (A)}\\
    CRNN &56.7&48.0&11.2&86.2&40.0&79.0&49.6&44.7&37.6&22.7&43.0&27.0&56.3&46.3&61.4&72.3\\
    \hline
    \textbf{Poster (P)}\\
    VGG16 &48.6&60.0&12.1&73.4&33.4&69.8&47.2&41.3&37.0&22.3&38.1&33.9&46.3&43.3&51.9&66.5\\
    \hline
    \textbf{Text (T)}\\
    Conv1D &62.5&34.4&24.7&64.8&54.3&73.8&50.3&64.6&50.4&31.5&43.2&70.6&61.5&52.8&57.8&70.4\\
    LSTM &64.8&44.5&25.6&70.1&63.4&78.0&63.3&70.8&63.2&32.6&47.1&75.2&66.5&58.9&63.8&73.8\\
    Bidirectional LSTM &63.7&42.5&31.2&69.3&58.1&76.7&57.9&66.4&61.3&30.7&52.3&76.2&63.2&57.7&63.2&73.5\\
    fastText &72.0&50.7&40.6&81.1&68.7&82.3&69.2&68.8&78.3&47.8&60.3&74.4&72.9&66.7&72.5&81.4\\
    fastText w/ Glove~\cite{pennington2014glove} &72.2&51.6&45.2&81.2&69.1&82.3&70.8&68.9&78.8&49.7&61.1&75.2&73.3&\textbf{67.7}&\textbf{72.8}&\textbf{81.7}\\
    \hline
    \textbf{Metadata (M)}\\
    XGBoost &61.5&76.8&35.4&74.8&36.7&82.7&83.7&53.7&62.3&22.8&31.4&33.4&50.9&54.3&62.9&73.7\\
    RandomForest &59.3&73.7&33.3&74.9&40.6&82.7&83.2&58.8&62.7&25.4&35.4&37.9&55.0&\textbf{55.6}&\textbf{63.9}&\textbf{73.7}\\
    \hline
    \textbf{Score Fusion}\\
    Video-Audio (VA) &69.0&90.8&26.1&88.6&49.0&82.6&74.8&63.8&49.0&34.4&49.8&51.1&70.8&61.5&70.3&78.8\\
    Vid-Aud-Poster (VAP) &68.8&92.5&27.4&88.5&48.9&82.6&74.8&63.7&49.5&34.3&50.1&50.3&70.7&61.7&70.4&78.8\\
    Vid-Aud-Post-Text (VAPT) &73.3&95.2&29.9&91.0&61.2&85.0&77.2&69.0&68.9&38.8&51.8&61.6&74.1&67.5&74.9&82.3\\
    Vid-Aud-Post-Text-Metad (VAPTM) &75.5&88.8&36.6&91.5&60.6&86.8&87.0&70.5&74.6&39.7&49.7&59.4&71.3&\textbf{68.6}&\textbf{75.3}&\textbf{82.5}\\
  \hline
\end{tabular}
}
\end{table*}

\vspace{0.04in}
\noindent{\bf Baselines:} Recurrent Neural Networks are a special type of neural networks used to represent sequences of arbitrary length, where an internal hidden representation allows it to carry information through time. RNNs were found to be effective in tasks such as image captioning and machine translation. But regular RNNs sometimes fail to capture the representation of longer sequences. Long Short-Term Memory networks~\cite{lstm-original} have shown to be a better alternative. We investigate whether LSTMs perform well in our task of movie genre classification since both video and text can be represented as sequences. For video, we additionally performed experiments using spatio-temporal feature learning using deep three-dimensional convolutional networks (C3D)~\cite{c3d}, and the Two-Stream Inflated 3D convolutional network (I3D)~\cite{i3d}, standard models used for action classification on videos. 

\vspace{0.04in}
\noindent{\bf Training:}  In experiments, we use $4,927$ video trailers after discarding movies which did not have any trailer or plot available, or for which its corresponding trailer is longer than $12$ minutes. We setup our training, validation and test splits by using three random non-overlapping sets containing 70\%, 10\% and 20\% of the data for each split respectively. This is 3449 samples for training, 491 samples for validation, and 987 samples for testing. In order to run fair comparisons we modify the RNNs and LSTMs by restricting their number of parameters (by limiting the size of hidden units and states) such that all the models compared have approximately the same representation power. The frequency of the genres is taken into consideration in the loss function; samples belonging to infrequent genres, like \texttt{biography}, are weighted more than commonly occurring classes like \texttt{drama}, with an attempt to mitigate label bias during classification. Across all the experiments, we use ADAM~\cite{adam} as the optimizer, decaying the learning rate by a factor of 0.001 after every epoch. Dropout~\cite{dropout} (with a drop rate of 0.5) was used as a regularization technique to control overfitting. The training was run for 100 epochs using a batch size of 32. All neural network parameters, with the exception of word embedding matrix weights were initialized using Xavier's initialization \cite{glorot}. 




\begin{figure*} 
\centering
\includegraphics[width=0.75\linewidth,height=3cm]{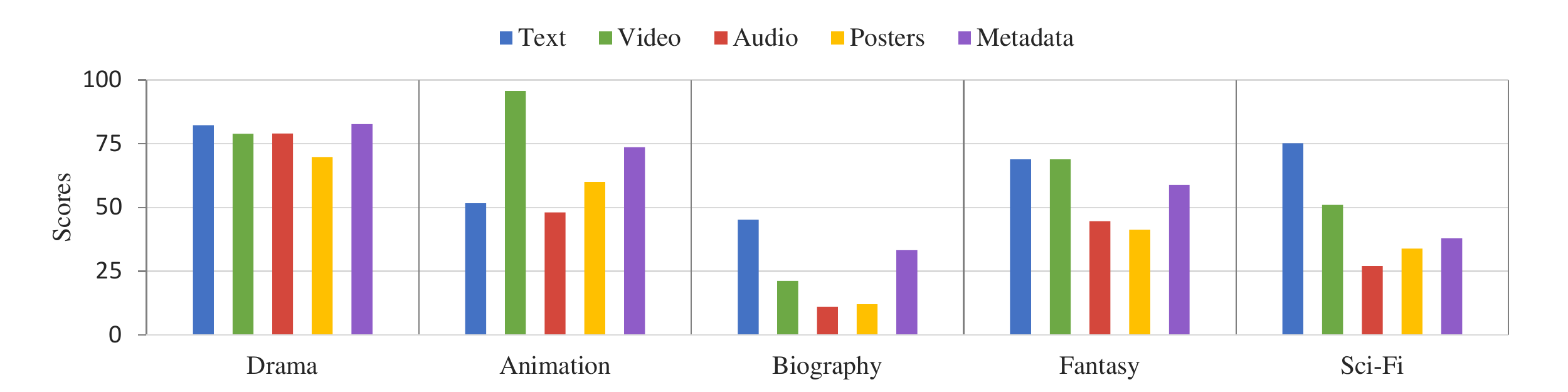}
\caption{Attention on different modalities for a select subset of genres.}
\label{fig:modal_attn}
\end{figure*}

\section{Results and Evaluation}
\label{sec:results}
We show in Table~\ref{tab:prc} the average precision for each method discussed in the paper, for each modality. The average precision was computed as the area under the precision recall curve of our predicted scores against the true labels. We also report in the last three columns the mean average precision ($mAP$) across categories (or macro-accuracy) which calculates the mean of the binary metrics giving equal weight to each class, the micro average precision ($\mu AP$) across samples which gives each sample-class pair an equal contribution to the overall metric, and the sample average precision ($sAP$) which does not calculate a per-class measure and instead calculates the metric over the true and predicted classes for each sample in the evaluation data, returning their weighted average. The first metric makes sure we are performing well across all categories, even for those that have less training samples or are more difficult to predict. The second metric ensures that we obtain overall good results across all samples in the dataset. The third metric can be considered as an atomistic evaluation in which each sample is measured independently from the whole set, and then gives us an averaged score.

As can be seen from Table~\ref{tab:prc}, LSTM models are not a good feature aggregation approach for this task, and while there is some improvement from using a single directional LSTM compared to a bidirectional LSTM, 
both fastVideo and fastText outperform all other methods.
When using LSTM models we found that choosing random clips of 16 or 49 continuous frames worked better than using 200 frames at once. At test time we divided the 200 frames we used for fastVideo into 12 clips of 16 frames and 4 clips of 49 frames, and we averaged the scores. By doing this, we are not forcing the LSTM to learn long-term dependencies at once, which yields to better results. Unlike LSTM models, when using fastVideo the larger amount of features yield better results.
We also found that standard models for action recognition on videos did not performed as well, only C3D outperformed the LSTM model. Our overall assumption is that due to the complexity of the long temporal semantics along with the dynamism present on our videos, complex and resource expensive models like C3D do not capture well spatio-temporal features. Flow-based methods such as I3D did not perform well, perhaps due to cuts along video trailers, and action interruptions on movie trailer scenes. 
This suggests that for holistic classification tasks, perhaps temporal ordering is not crucial, or that training LSTM models that take advantage of temporal dynamics requires a lot more training data.
In addition, fastText and fastVideo are considerably faster in both training and testing. Morever we show in table~\ref{table:ucf101} that fastVideo while cleary not state of the art on UCF101 does outperform similar competing approaches such as LSTM encodings or C3D but does not outperform flow-based methods such as I3D, clearly showing that in these scenarios flow-based features are still crucial.


\begin{table}[b]

  \centering
  \small
\caption{Mean Average Precision Scores on UCF101.}
\label{table:ucf101}
  \vspace{0.02in}
  \setlength{\tabcolsep}{12pt}
  {
  \label{tab:ucf101}
  \begin{tabular}{l|c}
    \hline
    &mAP\\
    \hline
    `Slow Fusion' spatio-temporal ConvNet \cite{slow-fusion} & 65.4 \\
    LSTM composite model (only RGB) \cite{lstm-rgb-composite} & 75.8 \\
    C3D (fc6) \cite{c3d} & 76.4 \\
    \hline
    iDT+C3D (fc6) \cite{c3d} & 86.7 \\
    Two-stream model \cite{two-stream} & 88.0 \\
    Two-Stream I3D \cite{i3d} & 98.0 \\
    \hline
    fastVideo - 16 Frames & 79.2 \\
    fastVideo - 200 Frames & 79.4 \\ 
    fastVideo - 49 Frames & 81.1 \\ 
    \hline
\end{tabular}
}
\end{table}

\begin{figure*}
    \centering
    \includegraphics[width=0.95\textwidth,keepaspectratio]{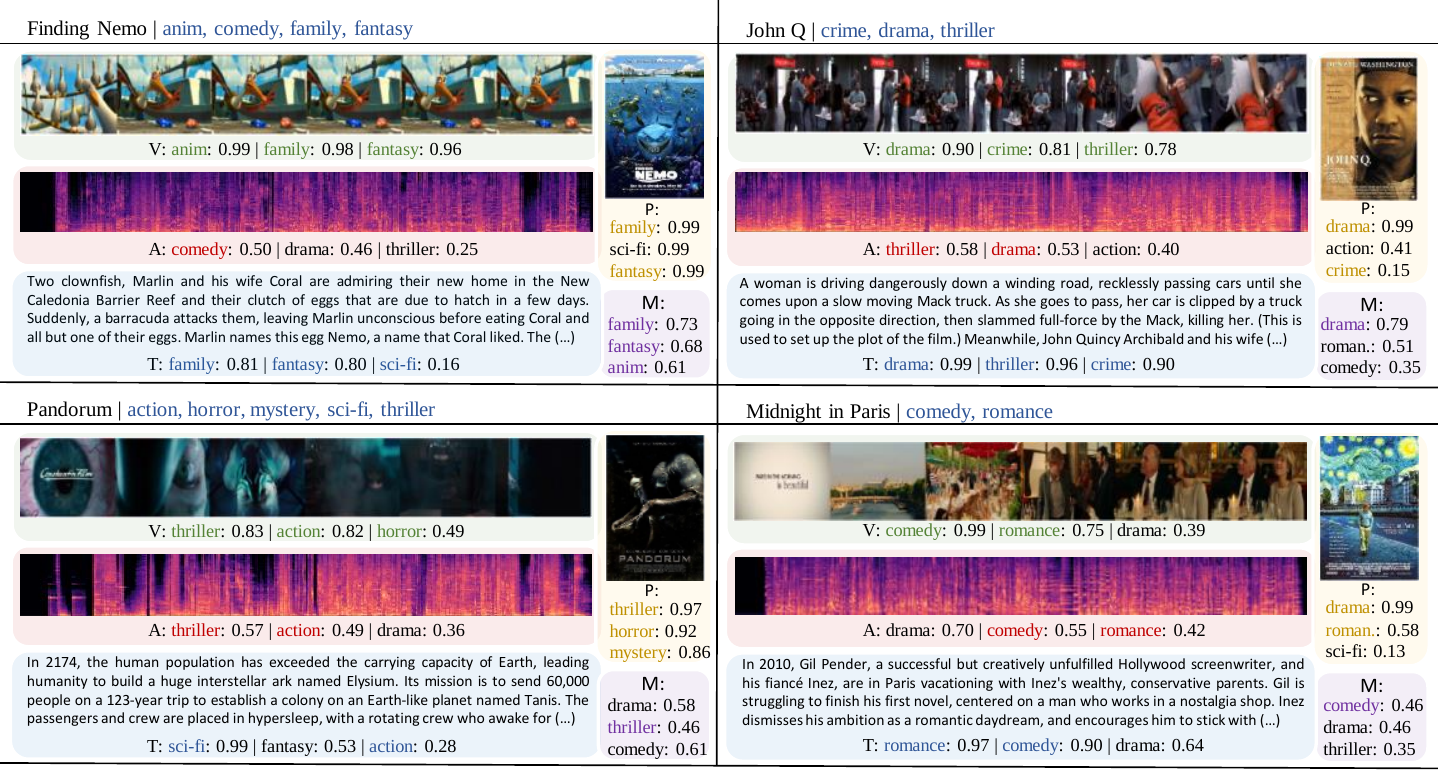}
    
    \centering\caption{Example predictions of the best performing models for each modality. We show some sample frames for the video trailers, and a spectrogram visualization for the audio.}
    \label{fig:qualitative}
\end{figure*}

Text-based representations outperformed video-based models, however more surprising is that text-based representations outperform our metadata-based model. For some categories such as \texttt{animation}, video representations were obviously more useful than text or metadata. Since the audio of the video trailer tries to capture the mood of the movie, fusing this representation with the video-based representation outperformed all LSTM-based models using only text representations. We also found that poster-based representation tended to be misleading, but combining the five modalities offers positive improvements. 
We show some prediction results for a few movies in Figure~\ref{fig:qualitative} for the five models, i.e., fastText-GloVe, fastVideo, audio-CRNN, posters-VGG16 and the metadata (Meta).

Finally, Figure~\ref{fig:modal_attn} shows the aggregated values of our \emph{modal attention} fusion scores represented by $\alpha_{i}$ for the $i^{th}$ modality. It can be observed that the modal 
attention weights corresponding to text are higher than the other modalities, which is consistent with individually observed results but we also observe clear differences across movie genres. 
For instance, video seems the best predictor for \texttt{animation} where the model gives more attention to the visual features. While trailer-based prediction does not outperform plot-based predictions, in combination it still improves the overall model accuracy for almost all categories.

\begin{table}[b]
\centering
\small
\centering\caption{Results for our budget prediction task.}
\vspace{0.02in}
  \label{tab:budget}

  \begin{tabular}{l|ccccc|c}
    \hline  
     &tier 1&tier 2&tier 3&tier 4&tier 5&$mAP$\\
    \hline
    Base &6.5&20.7&32.9&33.3&6.4&20.0\\
    \hline
    I3D Flows &8.2&21.2&29.9&39.8&20.3&23.9\\
    I3D RGB &16.7&30.4&38.5&47.0&28.6&32.2\\
    I3D~\cite{i3d}&13.9&29.0&36.4&46.1&32.7&31.6\\
    \hline
    fastVideo &19.6&17.1&34.8&36.8&55.9&32.9\\
    Audio &25.9&34.5&35.0&47.7&35.5&35.7\\
    fastText &19.4&36.5&37.6&46.6&40.6&36.1\\
    \hline
    Metadata &90.1&93.8&89.9&93.7&87.7&91.0\\
  \hline
\end{tabular}
\par
\vspace{0.02in}
\end{table}

\subsection{Automatic Budget Estimation}
We explored an additional task to examine if these modalities could help predict other characteristics in our movie dataset. We chose to predict the budget of each movie due to its possible relation to the quality of the video production, assuming that the number of video cuts and similar photography in the trailers could be related to the amount of money invested in a movie. We show in table \ref{tab:budget} results on the previous evaluated modalities in this particular task. Since the movies being evaluated are highly diverse, we decided to define ranges of budget expenses into 5 tiers as follows: \emph{tier-1} (\$218 to \$890K), \emph{tier-2} (\$900K to \$4.8M), \emph{tier-3} (\$4.9M to \$19.4M), \emph{tier-4} (\$19.5M to \$71.5M), and \emph{tier-5} (\$72M to \$300M). Unsurprisingly, metadata features such as \emph{number of critic reviews} or \emph{number of facebook likes} make this modality to surpass all others. Also, as opposed to our original assumption, audio and text modalities yield the better results. We also found out that there is a high variability in the correlation between genres and budget tiers, for example we found that tier 4 and 5 have the largest number of \emph{fantasy} and  \emph{action} movies, but other genres that one can intuitively relate to high budget movies such as \emph{sci-fi} is spread among all tiers. 

\subsection{Human-based vs Content-based Predictions}
In general, predicting a movie genre based on a short video or using a short text description seems to be a non-trivial task. Therefore we wanted to compare the human performance using only three modalities. We selected 100 movies from our test dataset and presented three different surveys for 40 different people. In the experiment, people had to indicate if they already watched the movie and then select one or more category genres for posters, text plots and 30 of the image frames we used for our video trailer evaluations. We used Amazon Mechanical Turk (AMT) for this experiment. In total, 629 people participated completing the surveys.
Table \ref{tab:human} shows the results of this evaluation. In general, people were better at identifying genres out of text and had more trouble with video frames identification. Moreover, people were more successful identifying movies they already watched by only looking at the posters. In contrast to our model, people were also good predicting genres using only the posters. This could indicate that since there is a weak correlation between the visual semantics of the posters and each genre, the model struggles predicting this task.
The relative error of the $mAP$ between the humans evaluations and our models for text is 0.029 and for video is 0.05. This indicates that humans performed slightly better than our models.

\begin{table}
\centering
\small
\centering\caption{Human evaluations. The last column W represents the percentage of people that recognized the movie for each modality tested. In parenthesis we show mean AP scores for people who reported not having seen movies they rated.}
\vspace{0.04in}
  \label{tab:human}
  \begin{tabular}{l|rrr|r}
    \hline  
     &$mAP$&$\mu AP$&$sAP$&\% W\\
    \hline
     Video&63.0 (62.0)&65.7&74.0&33.52\\
     Poster&71.7 (69.2)&75.9&84.6&50.45\\
     Text&72.7 (69.7)&70.9&79.4&35.77\\
  \hline
     VAPTM & 68.6&75.3&82.5&-\\
  \hline
\end{tabular}
\par
\vspace{-0.12in}
\end{table}

\vspace{-0.1in}
\section{Conclusions}
\vspace{-0.1in}
We propose the first multimodal dataset that includes movie trailers, with corresponding movie plots, movie posters, and associated metadata for $5,000$ movies. We also plan to release pre-trained encodings for both the video trailers: fastVideo, movie plots: fastText, and movie posters. We found that despite their simplicity, fastText and fastVideo are better suited for a holistic classification task than LSTM-based representations. We also revisit the classical question in multimedia of whether content matters~\cite{slaney2011web} with a fresh look in the era of deep learning. Our multimodal fusion experiment results show that deep learning models can be designed to provide better content for consumers without relying solely on metadata or user provided text. Given the enormous interest in both the vision community to study novel video representations for actions, and in the natural language processing community to study representations for movie plots, we hope to set a novel benchmark for joint tasks in this domain.

{\small
\bibliographystyle{ieee}
\bibliography{egpaper}
}

\end{document}